%% file: main.tex
\documentclass[fleqn,12pt]{wlscirep}
\usepackage[utf8]{inputenc}
\usepackage[T1]{fontenc}
\usepackage{xcolor}
\usepackage{xurl}
\usepackage{todonotes}
\usepackage{multirow}
\usepackage{adjustbox}
\usepackage{makecell}
\usepackage{array}     
\usepackage[table]{xcolor}
\usepackage[normalem]{ulem}
\usepackage{comment}
\useunder{\uline}{\ul}{}
\usepackage{lineno}
\usepackage{pifont}

\title{MedPMC: A Systematic Framework for Scaling High-Fidelity Medical Multimodal Data for Foundation Models}

\author[1]{Hyunjae~Kim}
\author[2]{Dain~Kim}
\author[3]{Pan~Xiao}
\author[1]{Serina~S.~Applebaum}
\author[1]{Younjoon~Chung}
\author[1]{Xuguang~Ai}
\author[4]{Yu~Yin}
\author[1]{Roy~Jiang}
\author[1]{Yuexi~Du}
\author[1]{Yawen~Wei}
\author[1]{Yiming~Kong}
\author[1]{Tuo~Guo}
\author[1]{Zhiyuan~Cao}
\author[1]{Mengmeng~Du}
\author[1]{Yuelei~Fu}
\author[5]{Yan~Hu}
\author[1]{Rui~Shi}
\author[1]{Gui~Yang}
\author[1]{Kevin~W.~Jin}
\author[1]{Yuntian~Liu}
\author[1]{Yuxuan~Tian}
\author[6]{Jonathan Marquez}
\author[1]{Zhen~Chen}
\author[7]{Sheng~Zhang}
\author[7]{Hoifung~Poon}
\author[1]{Hua~Xu}
\author[2]{Jaewoo~Kang}
\author[1,*]{Qingyu~Chen}

\affil[1]{Yale University, New Haven, CT, USA}
\affil[2]{Korea University, Seoul, South Korea}
\affil[3]{Washington University in St. Louis, St. Louis, MO, USA}
\affil[4]{The University of Queensland, Brisbane, QLD, Australia}
\affil[5]{The University of Texas Health Science Center at Houston, Houston, TX, USA}
\affil[6]{University of Washington, Seattle, WA, USA}
\affil[7]{Microsoft Research, Redmond, WA, USA}
\affil[*]{qingyu.chen@yale.edu}

\input{sections/abstract}

\begin{document}

\flushbottom
\maketitle
%
%
\thispagestyle{empty}

\newcommand{\draftcomment}[3]{%
  \todo[inline,color=#2!20]{\textbf{#1:}~#3}%
}



\input{sections/introduction}

\input{sections/results}

\input{sections/discussion}

\input{sections/methods}

\bibliography{references}

\section*{Acknowledgements}
This study is supported by the National Institutes of Health National Library of Medicine under Award Number R01LM014604 and R00LM014024.

\section*{Author Contributions Statement}
H.K. and Q.C. conceived and designed the study. 
H.K. and D.K. curated the pretraining datasets. 
H.K., D.K., P.X., and Y.Y. developed the data curation pipeline. 
H.K., X.A., Y.W., and Y.K.  configured the manual annotation interface and measured annotator disagreement. 
S.S.A., Y.W., Y.K., T.G., Z. Cao., M.D., Y.F., Y.H., R.S., G.Y., K.W.J., Y.L., and Y.T. contributed to data annotation for pipeline evaluation. 
P.X., S.S.A., R.J., Y.D., and J.M. conducted manual evaluation of the finalized dataset. 
H.K. and Y.C. performed CLIP model training and evaluation. 
H.K. performed MLLM training and evaluation. 
H.K. and Q.C. drafted the manuscript. 
Z. Chen., S.Z., H.P., H.X., and J.K. provided critical feedback on the manuscript and study design. 
All authors read and approved the final version of the manuscript.

\section*{Data Availability}

The MedPMC corpus, component-level benchmark resources, pretrained checkpoints, and associated metadata generated in this study are publicly available through the MedPMC collection on Hugging Face: \url{https://huggingface.co/collections/Yale-BIDS-Chen/medpmc}. Because MedPMC is derived from articles with heterogeneous open licenses, each released record includes source identifiers and license metadata to support license-aware filtering, redistribution, and downstream model development. Articles with licenses that prohibit redistribution of derived resources were excluded from the released corpus. Users are responsible for following the license terms associated with the original source articles. In particular, some records are derived from articles carrying CC BY-NC or CC BY-NC-SA licenses, which restrict use to non-commercial purposes; users intending commercial applications should verify licensing compatibility and apply license-aware filtering before downstream use.

Each MedPMC release is versioned by PMC article cutoff date, source-license filters, curation-pipeline component checkpoints, and processing configuration, enabling users to reproduce a specific corpus snapshot or apply future pipeline updates to newly available literature.

The Yale New Haven Health System dermatology patient data used for clinical transfer evaluation are not publicly available because they contain patient-derived clinical data and are subject to institutional privacy, governance, and IRB restrictions. Aggregate evaluation results and processing procedures are described in the manuscript.

\section*{Code Availability}
The code used for data curation and pipeline evaluation is publicly available at:
\url{https://github.com/Yale-BIDS-Chen-Lab/MedPMC}.

\section*{Competing Interests}
The authors declare no competing interests.

\clearpage 

\appendix
\renewcommand{\tablename}{Extended Data Table}
\renewcommand{\figurename}{Extended Data Figure}
\setcounter{table}{0}
\setcounter{figure}{0}

\input{sections/extended_data}

\end{document}

%% file: sections/abstract.tex
\begin{abstract}

Medicine is inherently multimodal, requiring clinicians to synthesize information across diverse data streams. Yet the development of multimodal foundation models is constrained by limited access to large-scale, high-quality clinical data. Although PubMed Central (PMC) offers a complementary source of expert-authored image-text data, existing PMC-derived resources remain limited in fidelity, reproducibility, and clinical validation. We introduce MedPMC, an automated, continuously updatable framework that transforms permissively licensed literature into high-fidelity infrastructure for medical multimodal models. Applied to 6.1 million PMC articles, MedPMC curated 11 million medical image-text pairs. Component evaluations showed strong performance for initial screening (F1 = 93.2), multi-panel figure detection (F1 = 96.5), figure separation (mAP = 89.8), caption separation and alignment (F1 = 81.4; ROUGE-L = 85.3), and medical figure classification (F1 = 96.5). Manual review by five annotators, three with medical training, found 95.3\% of MedPMC images medically relevant, versus 19.7\% in a prior PMC-derived dataset. Across 26 benchmarks spanning 11 specialties, a MedPMC-trained CLIP-style model improved average zero-shot AUC by 7.1 percentage points over the strongest architecture-matched biomedical CLIP baseline despite using fewer than half as many image-text pairs. As the vision encoder in a multimodal large language model, it improved medical visual question-answering by 1.9 and 16.9 percentage points across two benchmarks. In 10,524 Yale New Haven Health System dermatology photographs, it improved morphology-to-image retrieval Recall@5 by 11.7 percentage points. These findings show that high-fidelity literature curation strengthens medical multimodal foundation models across benchmark and clinical settings. We publicly release the framework, corpus, benchmarks, and pretrained models.

\end{abstract}

%% file: sections/introduction.tex
\section{Introduction}
Medicine is multimodal in nature~\cite{acosta2022multimodal,moor2023foundation}.
Triage, diagnosis, prognosis, and treatment planning routinely require clinicians to integrate images, text, laboratory values, and other complementary data streams~\cite{cui2023deep,huang2020fusion}.
In principle, medical foundation models could unlock this potential by providing reusable backbones that generalize across medical modalities and can be adapted to local clinical settings with limited labeled data~\cite{he2024foundation,rajendran2024learning}.
In practice, however, current medical AI systems are still largely developed and evaluated within single-modality, single-task settings rather than on the heterogeneous evidence used in clinical care~\cite{schouten2025navigating,zhang2024challenges}.
This limitation stems not from model architecture alone, but from the absence of large-scale, high-quality multimodal data that are publicly accessible for training and evaluation, shareable across institutions, and suitable for reproducible research~\cite{zhang2024challenges,schouten2025navigating,acevedo2020dataset,huang2025systematic}.
Unlike curated benchmark datasets, real-world medical data are fragmented across devices and health systems; constrained by privacy, licensing, and governance requirements; and costly to annotate at scale~\cite{koul2025synthetic,liu2022real,mandreoli2022real}.
Although pioneering efforts have been made to make clinical datasets publicly accessible~\cite{irvin2019chexpert,huang2023inspect,ai2024ai}, these resources remain much smaller and narrower than the data used to train general-domain foundation models.
These barriers have led to two persistent challenges: a benchmark-to-bedside gap, in which models that perform well on public benchmarks may not deliver comparable utility in real clinical settings~\cite{sokol2025artificial,jiang2026multimodal,bedi2026holistic}; and fragmented, institution-specific curation efforts that limit reproducibility, scalability, and reuse~\cite{hughes2023addressing,huang2025open}.

One practical strategy is to leverage publicly available biomedical resources---including web-scale data, biomedical literature, and medical textbooks---to develop public foundation model backbones that are shareable, reproducible, and ultimately adaptable to clinical settings with less reliance on locally curated data~\cite{cao2025development,huang2025systematic,zhang2023pmc,chen2025compound}.
Among these sources, PubMed Central (PMC) has emerged as one of the most promising and widely used resources~\cite{huang2025systematic,cao2025development}.
PMC hosts millions of expert-authored biomedical articles that pair diverse medical visuals with explanatory text across a broad range of specialties, and it continues to expand rapidly as new biomedical knowledge is published~\cite{pmc_intro}.
Prior studies have leveraged PMC figures and associated text to develop medical vision-language models~\cite{lin2023pmc,eslami2023pubmedclip,wang2022medclip} and, more recently, generative multimodal large language models (MLLMs)~\cite{li2023llava,qin2026volmo}.

However, biomedical literature is created to communicate scientific findings, not to provide ready-to-use training data for multimodal foundation models.
Repurposing it into high-fidelity multimodal training data poses substantial challenges.
First, PMC contains a large fraction of visual content that is not directly useful for learning clinically relevant visual representations, including graphs, charts, schematics, flow diagrams, and molecular illustrations.
Second, figures in the biomedical literature are often compound multi-panel images; without accurate decomposition of subfigures and alignment with their corresponding subcaptions, the image-text correspondence required for effective multimodal learning is substantially degraded.
Finally, the continuous expansion of biomedical literature requires curation frameworks capable of dynamically incorporating newly published content. 
To date, existing PMC-derived datasets have not comprehensively addressed these challenges. For example, in a recent 24-million image-text pair dataset~\cite{lozano2025biomedica}, our analysis revealed that 80.3\% of the images were non-medical. 
Moreover, compound figures accounted for 61\% of the data but were not decomposed into individual panels. 
Furthermore, because most existing datasets are released as static snapshots, they become progressively outdated and fail to continuously integrate newly published knowledge.
These limitations are further detailed in \textbf{Extended Data Table~\ref{table:pmc_dataset_comparison}} and the Discussion section.

To address this, we developed MedPMC, an automated and systematic framework for curating high-fidelity medical multimodal data from permissively licensed biomedical literature.
MedPMC extends beyond a static dataset and instead provides reusable, continuously evolving data infrastructure for developing medical multimodal foundation models, including contrastive vision-language encoders and MLLMs.
Our contributions operate at three levels.
At the data resource level, MedPMC yields 11 million medical image-text pairs from 6.1 million PMC articles published through June 2024.
The framework consists of five sequential stages---initial screening, multi-panel figure detection, multi-panel figure separation, caption separation and alignment, and medical figure classification---each handled by a purpose-built model to improve the fidelity of the resulting training data (\textbf{Fig.~\ref{figure:study_design}a}).
Manual evaluation by five annotators, including three medical experts, showed that 95.3\% of curated images were medically relevant.
The resulting corpus spans a broad range of specialties, including pathology, radiology, ophthalmology, dermatology, and endoscopy, and includes clinically important modalities that are less commonly represented at scale in public AI datasets, such as slit-lamp photography, skin photography, gross pathology photography, endoscopy, angiography, PET/SPECT, and fMRI.
Importantly, MedPMC is designed for sustainable expansion rather than one-time dataset construction.
Because each processing stage is modular and supported by a dedicated benchmark, individual components can be systematically evaluated, replaced, and refined as improved methods become available.
This design allows MedPMC to continuously incorporate newly published biomedical literature, and the framework has identified more than one million new medically relevant image-text pairs per year since 2020.

\input{figures/study_design}

At the foundation model level, we trained MedPMC-CLIP, a CLIP-style vision-language model~\cite{radford2021learning}, as one representative instantiation to test whether high-fidelity MedPMC curation translates into stronger medical visual representations (\textbf{Fig.~\ref{figure:study_design}b}).
MedPMC-CLIP achieves state-of-the-art performance across 26 public medical benchmarks spanning 11 specialties, covering clinically relevant tasks such as pneumonia detection, diabetic retinopathy grading, lymph node metastasis detection, and colon cancer detection.
The model improves average AUC by 7.1 percentage points over BMC-CLIP~\cite{lozano2025biomedica}, a strong biomedical vision-language baseline, despite being trained on less than half the training data.
These gains are consistent across specialties, demonstrating that high-fidelity curation produces visual representations that generalize broadly across medical domains.
To further assess whether these representations can support generative multimodal systems, we conducted a controlled experiment following the training strategy introduced by LLaVA-Med~\cite{li2023llava}.
With all other components held fixed, replacing only the vision encoder with MedPMC-CLIP improves performance by 1.9 and 16.9 percentage points on two multimodal medical question-answering (QA) benchmarks, respectively.

At the clinical validation level, beyond standard benchmark evaluation, we further assessed transfer to downstream clinical applications.
Using 10,524 dermatology patient images derived from a multi-site, hospital-based dermatology consultation cohort within the Yale New Haven Health System (YNHHS), we evaluated a morphology-to-image retrieval task in which textual descriptions of clinical findings were used to retrieve visually similar patient images, reflecting a clinically meaningful image-retrieval use case relevant to diagnostic decision-making~\cite{karthik2025content,zhou2022cross} (\textbf{Fig.~\ref{figure:study_design}c}).
MedPMC-CLIP improves Recall@5 by 11.7 percentage points over BMC-CLIP, demonstrating that MedPMC-derived supervision can transfer effectively to clinically realistic settings.
This translational benefit is further supported by embedding-space analysis, which shows that MedPMC better captures the visual distribution of real patient images from YNHHS, with broader coverage and closer alignment than existing dermatology datasets.

Collectively, these results position MedPMC as scalable data infrastructure for medical multimodal foundation models, linking high-fidelity literature-derived data curation to stronger visual representations, improved multimodal reasoning, and broader generalization across medical domains and downstream tasks.
To support reuse and reproducibility, we publicly release the MedPMC framework, curated corpus, benchmark resources, and pretrained model checkpoints (see Data Availability and Code Availability).

%% file: figures/study_design.tex
\begin{figure}[t!]
\centering
\includegraphics[width=\linewidth]{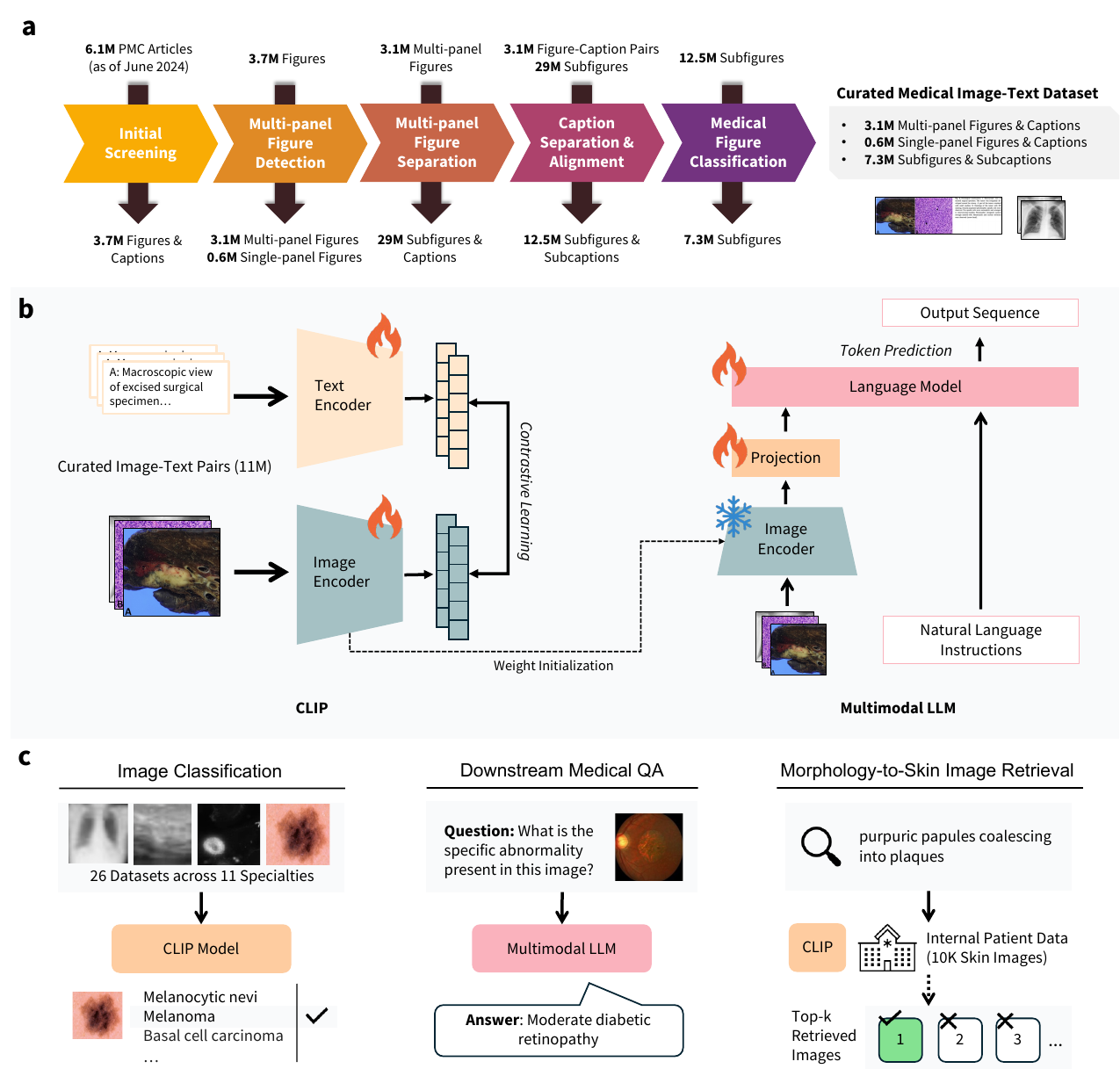}
\caption{
\textbf{MedPMC framework for dataset curation and its application to multimodal model training.}
\textbf{a}, Overview of the proposed five-step framework for constructing a large-scale medical image-text dataset from PMC articles. 
This pipeline yields 11M medical image-text pairs with improved data quality.
\textbf{b}, Model development pipeline. A CLIP-style vision-language model is first trained on the curated dataset via contrastive learning. 
The resulting vision encoder is then used to initialize a large vision-language model following the LLaVA-Med training framework, enabling multimodal instruction tuning for downstream medical reasoning tasks.
\textbf{c}, Evaluation across three complementary settings. The trained models are assessed through (i) zero-shot medical image classification across 26 datasets spanning 11 specialties, (ii) downstream medical QA with an MLLM, and (iii) morphology-to-skin image retrieval using internal patient data.
}
\label{figure:study_design}
\end{figure}

%% file: sections/results.tex
\section{Results}

\subsection{Pipeline Design and Data Curation}
\textbf{Fig.~\ref{figure:study_design}a} illustrates the overall curation framework.
The framework was designed to transform permissively licensed biomedical literature into high-fidelity medical image-text data through a sequence of automated, modular processing stages.
Here, we summarize the purpose of each stage and the resulting data outputs; detailed methodological descriptions are provided in the Methods section.

\paragraph{Definition of medical images}
PMC contains a broad spectrum of biomedical visual content, ranging from clinical imaging studies to basic science experiments, molecular biology figures, statistical plots, and conceptual diagrams.
In this study, we use ``medical images'' as an operational umbrella term to refer to visual content relevant to clinical medicine, human health, or translational biomedical research.
Our goal was to curate visual evidence most relevant for clinical and translational medical AI applications, including image interpretation, diagnosis, disease assessment, treatment planning, healthcare workflows, and medical research involving human-derived specimens~\cite{acosta2022multimodal,rao2025multimodal}.
We therefore prioritized image types routinely encountered in clinical practice, as well as human-relevant biomedical imaging modalities~\cite{mcauliffe2001medical,ogut2025artificial}, including radiology, ophthalmology, pathology, microscopy, endoscopy, dermatology, physiological signals, surgical photographs, and other forms of human clinical photography.
In contrast, we excluded visual content primarily depicting non-human organisms, basic cellular or molecular biology experiments without direct human medical or translational relevance, scientific plots, graphs, tables, conceptual diagrams, educational illustrations, and other materials outside this scope.

\paragraph{Training data construction}

The five-stage curation pipeline relies on purpose-built models for each component. We first leveraged existing labeled benchmarks, including ImageCLEF 2016~\cite{GSB2016}, MedICaT~\cite{subramanian2020medicat}, and DocFigure~\cite{jobin2019docfigure}, to train these models. However, these resources did not fully cover all stages of the curation process and were insufficient for several tasks. 
For instance, no public benchmark was available for the initial screening stage, while only 1,600 labeled examples were available for caption separation and alignment despite the complexity of the task.
We therefore supplemented existing benchmarks with PMC-derived synthetic annotations. From 39,360 sampled figures, we generated 99,321 labeled examples for the initial screening, multi-panel figure detection, and caption separation and alignment stages, complementing the 50,995 manually labeled samples available from existing benchmarks (\textbf{Extended Data Table~\ref{table:pipeline_train_test_data}}).
This expanded training set substantially improved performance across the affected pipeline stages. 
For example, augmenting with PMC-derived synthetic data improved the validation F1 score for the caption separation and alignment task by 3.82 percentage points over using the labeled data alone (see \textbf{Supplementary Information, Section 1} for details). 
Importantly, the final reported performance of each curation component was assessed on held-out real instances with manual labels.

\paragraph{Initial screening}
The initial screening stage was designed to identify candidate medical figures while avoiding the need to download all PMC images, which is time-consuming and storage-intensive at scale.
We first collected all open-access articles in PMC available as of June 2024, yielding 6,106,189 articles.
Among these, 5,099,175 articles carried permissive licenses that allowed downstream use and redistribution, including CC BY (3,940,226), CC BY-NC (742,418), CC BY-NC-SA (270,631), CC0 (144,680), and CC BY-SA (1,220).
We excluded articles licensed under CC BY-NC-ND (592,009) and CC BY-ND (8,188), as these licenses prohibit redistribution of derived resources, as well as records labeled as NO-CC CODE (406,817), which typically indicate missing, custom, or non-machine-readable license terms.
We then trained a text classifier that used figure captions and associated inline reference text to distinguish candidate medical figures from non-medical figures.
Applied to the 5.1 million permissively licensed articles, this stage retained 3.7 million candidate image-text pairs for downstream processing.

\paragraph{Multi-panel figure detection}
After initial screening, we downloaded the retained candidate images and determined whether each image was single-panel or multi-panel.
Single-panel images could be directly retained for downstream curation, whereas multi-panel figures required further decomposition and caption alignment.
We therefore trained an image classifier to distinguish single-panel from multi-panel figures.
This stage identified 3.1 million multi-panel figures.

\paragraph{Multi-panel figure separation}
We next decomposed the identified multi-panel figures into individual subfigures.
A figure separation model was trained to predict bounding boxes for panel-level image crops.
In total, 3.1 million multi-panel figures were decomposed into an average of 9.2 sub-panels, yielding 29 million subfigures.
This high panel density is consistent with the layouts commonly observed in biomedical articles, including microscopy grids, multi-condition comparisons, and multi-modality clinical figures.

\paragraph{Caption separation and alignment}
After figure separation, we processed the corresponding full captions to construct panel-level image-text pairs.
This stage involved two linked tasks: caption separation, which decomposes a full figure caption into subcaptions, and subfigure-subcaption alignment, which associates each subcaption with its corresponding subfigure.
Rather than performing these tasks sequentially~\cite{subramanian2020medicat,lin2023pmc}, we addressed them jointly using an MLLM.
The model takes as input the original multi-panel figure, the extracted subfigures, and the full caption, and directly generates a list of subcaptions corresponding to the input subfigures.
Generated subcaptions were aligned to subfigures according to their order.
To prioritize pair fidelity, we removed samples in which the number of generated subcaptions did not match the number of input subfigures.
These cases typically involved figures with many subpanels or ambiguous semantic boundaries, where reliable panel-level correspondence is inherently difficult to establish.
Although this filtering step reduced the total number of retained pairs, it improved the fidelity of the resulting image-text pairs.
After this process, we obtained 12.5 million high-quality subfigure-subcaption pairs.

\paragraph{Medical figure classification}
Because multi-panel figures often contain mixtures of medical and non-medical visual elements, such as charts, diagrams, or experimental schematics, we further filtered the decomposed subfigures.
We trained an image classifier to determine whether each subfigure was medically relevant.
After this final filtering step, we retained 7.3 million medical subfigure-subcaption pairs.

\subsection{Stage-level Benchmark and Pipeline Performance}

\paragraph{Benchmark curation}
A central challenge in large-scale medical multimodal data curation is that errors introduced at intermediate stages can propagate through the pipeline and degrade the final image-text pairs~\cite{adiba2026multimodal,wornow2023shaky}.
To systematically evaluate and optimize each component of MedPMC, we constructed a benchmark suite for component-level assessment across the full curation pipeline.
This benchmark enables direct comparison between our purpose-built models and prior approaches used in PMC-derived dataset construction, while also providing a standardized basis for future refinement of individual pipeline components.
The benchmark integrates three complementary data sources: (i) repurposed existing labeled benchmarks, including ImageCLEF 2016, MedICaT, and DocFigure; (ii) newly constructed PMC-derived synthetic annotations; and (iii) newly annotated samples curated by human annotators (see \textbf{Extended Data Table~\ref{table:pipeline_train_test_data}} for dataset statistics and composition).
Similar to our strategy for supplementing pipeline training data, we added PMC-derived synthetic annotations and newly annotated samples because existing benchmarks did not fully cover all pipeline stages or provide sufficient examples for robust evaluation, particularly for the caption separation and alignment task, which requires structured caption parsing and subfigure-level alignment rather than simple image classification and is therefore substantially more difficult and costly to annotate.
The validation set was designed to assess in-distribution performance relative to the pipeline training data and included 11,357 labeled examples and 19,562 synthetic examples across all stages. The held-out test set comprised 24,068 labeled samples from existing benchmarks and 2,010 newly annotated examples.
Additional details on benchmark construction and annotation procedures are described in the Methods section.

\paragraph{Results}
Using this benchmark, we evaluated each component of the MedPMC pipeline against baseline methods, including previous PMC-based data curation approaches (e.g., MedICaT and PMC-OA~\cite{lin2023pmc}), task-specific state-of-the-art models, and component-specific ablations. 
Detailed descriptions of all baseline methods and complete component-level results are provided in \textbf{Supplementary Information, Section 1}.
For the initial screening stage, we compared multiple biomedical language models as well as different textual inputs, including captions alone and captions combined with inline text. 
The PubMedBERT classifier~\cite{gu2021domain} using both captions and inline text achieved the best performance (F1 = 93.2), substantially outperforming the keyword-matching baseline used in MedICaT (F1 = 61.7).
For multi-panel figure detection, we compared text-based approaches, multiple vision model architectures, and the effect of synthetic training data. 
The Vision Transformer~\cite{dosovitskiy2021image} trained using both labeled and synthetic training data achieved the best performance (F1 = 96.5).
For figure separation, the previous state-of-the-art model, YOLO-OCR-D~\cite{meng2024yolo}, reported strong performance (mAP = 90.2), but it is not publicly available. 
We therefore trained a YOLOv10-based detector~\cite{wang2024yolov10} using the ImageCLEF dataset, achieving comparable performance (mAP = 89.8) while enabling a fully reproducible pipeline.
For caption separation and alignment, we compared the rule-based caption parsing and CLIP-based alignment strategy used by PMC-OA, GPT-4 Turbo (hereafter referred to as GPT-4T) in a zero-shot setting~\cite{hurst2024gpt}, and supervised InternVL-2.5-4B models~\cite{chen2024expanding} trained using manually annotated MedICaT data with or without synthetic augmentation. 
The supervised InternVL-2.5-4B model achieved the best performance (F1 = 81.4, ROUGE-L = 85.3), substantially outperforming both the PMC-OA pipeline (F1 = 48.5, ROUGE-L = 44.4) and GPT-4T zero-shot (F1 = 75.2, ROUGE-L = 76.9).
Finally, for medical figure classification, we compared four vision models, with the Vision Transformer again achieving the best performance (F1 = 96.5).

\subsection{Dataset Composition, Coverage, and Scalability}

\input{figures/data_analysis}

The final MedPMC dataset comprises 3.1 million multi-panel figure-caption pairs, 0.6 million single-panel image-caption pairs, and 7.3 million subfigure-subcaption pairs extracted from multi-panel figures, totaling approximately 11 million medical image-text pairs.
Although MedPMC is smaller than broader PMC-derived collections such as PMC-15M and BIOMEDICA~\cite{lozano2025biomedica}, those resources are not restricted to medical images and include a large proportion of non-medical content.
To quantify this difference in dataset composition and assess the clinical relevance of the curated images, we conducted a manual evaluation with five annotators, including medical experts.
We analyzed 2,906 samples from MedPMC and 432 samples from BIOMEDICA for comparison.
Multi-panel images were excluded from this analysis because assigning a single image category to a compound figure is inherently ambiguous.

As shown in \textbf{Fig.~\ref{figure:data_analysis}a}, BIOMEDICA contained a substantial proportion of non-medical images (80.3\%), with statistical figures, graphs, and charts accounting for 42\% of sampled images.
In contrast, MedPMC was predominantly composed of medical images, with only 4.7\% categorized as non-medical.
A substantial amount of images covered clinically relevant modalities routinely encountered in clinical practice, including radiology, ophthalmology, pathology, endoscopy, dermatology, physiological signals, surgical imaging, and other forms of clinical photography. Additionally, non-pathology microscopy provides complementary coverage of medical imaging beyond routine clinical practice.
\textbf{Fig.~\ref{figure:data_analysis}b} further breaks down the radiology subset into finer-grained modalities.
MRI (31.7\%), CT (23.4\%), and X-ray (15.1\%) constituted the largest radiology categories, while ultrasound, PET/SPECT, fMRI, angiography, and mammography were also represented.
Detailed distributions across additional domains are provided in \textbf{Extended Data Table~\ref{table:data_analysis}}.
\textbf{Fig.~\ref{figure:data_analysis}c} illustrates the scalability of the MedPMC framework by quantifying the annual number of publications and corresponding figure-caption pairs processed by the pipeline.
The volume of biomedical publications increased steadily over time, accompanied by a proportional increase in extracted image-text pairs.
Since 2020, the framework has identified more than one million image-text pairs per year, demonstrating its capacity to continuously incorporate a vast amount of newly published biomedical literature.

\input{figures/clip_results}

\subsection{Foundation Model Evaluation and Clinical Transfer Validation}

To evaluate whether high-fidelity MedPMC curation translates into stronger medical visual representations and downstream multimodal reasoning, we evaluated MedPMC through two representative model instantiations.
First, we trained MedPMC-CLIP, a CLIP-style vision-language model, following the exact training protocol of BMC-CLIP~\cite{lozano2025biomedica}, including identical architecture, initialization, configuration, hyperparameters, and training schedule.
Specifically, we initialized MedPMC-CLIP from the same OpenCLIP ViT-L/14 weights used by BMC-CLIP and replaced only the BIOMEDICA 24M training corpus with MedPMC.
This design enabled a head-to-head comparison that isolated the effect of dataset quality.
Second, we examined whether the visual representation learned by MedPMC-CLIP could serve as a stronger visual backbone for generative multimodal systems.
We adopted the LLaVA-Med training pipeline~\cite{li2023llava}, which builds on LLaVA v1.5~\cite{liu2023visual} and performs the visual alignment and instruction tuning stages.
While keeping the training procedure unchanged, we replaced only the original vision encoder with MedPMC-CLIP.
This controlled design tested whether improved visual pretraining transfers to downstream medical visual question answering.
Detailed results with 95\% confidence intervals are provided in \textbf{Supplementary Information, Section 2}.

\paragraph{MedPMC-CLIP improves performance across medical specialties}

We evaluated MedPMC-CLIP on image classification across 26 public medical benchmarks spanning 11 specialties.
\textbf{Extended Data Table~\ref{table:list_of_benchmarks}} lists the benchmarks, specialties, and task definitions included in the evaluation.
Together, these benchmarks cover disease detection and classification, severity grading, tissue classification, anatomical structure recognition, cellular phenotyping, and morphological or ultrastructural identification.
Clinically relevant tasks include pneumonia detection, breast malignancy classification, diabetic retinopathy grading, colon adenocarcinoma detection, lymph node metastasis detection, heart failure classification, and skin lesion classification.
\textbf{Fig.~\ref{figure:main_results}a} summarizes performance across the 26 benchmarks.
For each benchmark, performance was first averaged within specialty, and the final score was computed by averaging across 11 specialties.
MedPMC-CLIP consistently outperformed baseline models under this evaluation.
Compared with BMC-CLIP, MedPMC-CLIP improved the specialty-macro average accuracy by 7.1 percentage points (95\% CI, 6.0--8.2), F1 score by 10.5 percentage points (95\% CI, 9.2--11.6), and AUC by 7.1 percentage points (95\% CI, 6.3--8.0; paired bootstrap with 10,000 replicates).
\textbf{Fig.~\ref{figure:main_results}b} presents performance by specialty.
MedPMC-CLIP outperformed BMC-CLIP in 10 of 11 specialties, indicating that the gains were not confined to a single domain but were broadly distributed across diverse medical fields.

\paragraph{MedPMC-CLIP strengthens downstream multimodal medical QA}

\textbf{Fig.~\ref{figure:main_results}c} shows that replacing the vision encoder with MedPMC-CLIP yielded an observed improvement of 1.9 percentage points on MMMU (95\% CI, $-4.1$ to 8.0 percentage points) and a substantially larger improvement of 16.9 percentage points on OmniMedVQA (95\% CI, 14.9--18.7 percentage points; paired bootstrap with 10,000 replicates).
This finding highlights the role of the vision encoder in medical multimodal reasoning and is consistent with prior studies showing that vision encoder choice~\cite{mckinzie2024mm1}, as well as domain-specific adaptation of the visual backbone~\cite{zambrano2025clinically}, can substantially affect downstream multimodal LLM performance.
The larger gain on OmniMedVQA was primarily driven by tasks in which visual recognition plays a central role. 
We analyzed performance according to the query categories provided by OmniMedVQA. Performance improved from 46.9\% to 75.2\% for modality recognition, corresponding to an increase of 28.3 percentage points (95\% CI, 24.2--32.4); from 34.7\% to 57.0\% for anatomy identification, an increase of 22.3 percentage points (95\% CI, 17.6--26.9); and from 26.0\% to 40.8\% for the ``other biological attributes'' category, an increase of 14.8 percentage points (95\% CI, 7.6--22.0). These findings suggest that a stronger medical vision encoder can directly improve the model's ability to interpret medical images.
Importantly, the improvement was not limited to visually straightforward recognition tasks. 
Performance on the combined ``disease diagnosis and lesion grading'' category increased from 34.2\% to 45.4\%, corresponding to an improvement of 11.1 percentage points (95\% CI, 8.5--13.8). 
In contrast, MMMU questions are often framed around patient presentations and require complex clinical reasoning; thus, although the point estimate favored the stronger vision encoder, its confidence interval included zero, suggesting that further gains may require more effective end-to-end instruction tuning.

\paragraph{MedPMC-CLIP improves morphology-guided retrieval of clinical dermatology images}
Beyond standard benchmarking, we conducted a clinical evaluation using 10,524 dermatology patient images derived from a multi-site, hospital-based dermatology consultation cohort within the Yale New Haven Health System (YNHHS).
Specifically, we evaluated MedPMC-CLIP in a clinically motivated morphology-to-skin-image retrieval task, in which textual morphology descriptions are used to retrieve relevant clinical images.
This setting serves as a probe of representation learning and reflects a clinically meaningful retrieval use case, as retrieving similar medical images has long been studied as a form of decision support and education in dermatology and broader medical imaging contexts~\cite{sadeghi2020using,gassner2023saliency,muller2004review,long2009content}.
As shown in \textbf{Fig.~\ref{figure:main_results}d}, MedPMC-CLIP outperformed BMC-CLIP, improving Recall@1 by 3.4 percentage points (95\% CI, 2.7--4.1), Recall@5 by 11.7 percentage points (95\% CI, 10.7--12.6), and Recall@10 by 10.7 percentage points (95\% CI, 9.8--11.6; paired bootstrap with 10,000 replicates).
These results indicate that improvements observed on public benchmarks transfer to a clinically realistic image retrieval setting using patient data.

\input{figures/visualization}

\paragraph{MedPMC dermatology images better match the visual distribution of clinical images}

Although MedPMC images are derived from the biomedical literature rather than routine clinical practice, MedPMC-CLIP showed improved performance on downstream clinical dermatology retrieval.
To better understand the factors driving this transferability, we examined whether dermatology images curated from MedPMC occupy a visual space similar to that of real-world clinical dermatology photographs.
Specifically, we compared MedPMC with three widely used public dermatology datasets containing skin photographs (Fitzpatrick17k~\cite{groh2021evaluating}, SCIN~\cite{ward2024creating}, and DermNet~\cite{dermnet}) and assessed their respective visual similarity to the YNHHS clinical cohort.
For each public dataset, we sampled 10,000 images; for MedPMC, we first identified skin photographs using the MedPMC-CLIP model and then randomly sampled 10,000 images from this curated subset.

In the DINOv2 embedding space, MedPMC demonstrated the greatest visual similarity to YNHHS clinical images among all evaluated external sources.
Quantitatively, YNHHS clinical images yielded the lowest median nearest-neighbor cosine distance to MedPMC (0.239), compared with 0.262 for Fitzpatrick17k, 0.270 for SCIN, and 0.299 for DermNet (\textbf{Fig.~\ref{figure:visualization}a}).
Consistently, when each YNHHS clinical image was assigned to the external source of its nearest neighbor, MedPMC accounted for the majority of assignments (51.5\%), followed by Fitzpatrick17k (21.2\%), SCIN (20.7\%), and DermNet (6.5\%) (\textbf{Fig.~\ref{figure:visualization}b}).
UMAP projections further revealed that while YNHHS images formed a broad visual manifold that reflects the inherent variability of clinical photography, MedPMC showed extensive overlap across this space.
In contrast, existing public datasets exhibited more uneven distributions, leaving substantial portions of the clinical manifold underrepresented (\textbf{Fig.~\ref{figure:visualization}c--f}). 
These findings indicate that MedPMC captures a diverse visual distribution that closely aligns with real-world clinical data, providing a potential explanation for its downstream transfer to clinical dermatology tasks.

%% file: figures/data_analysis.tex
\begin{figure}[t!]
\centering
\includegraphics[width=\linewidth]{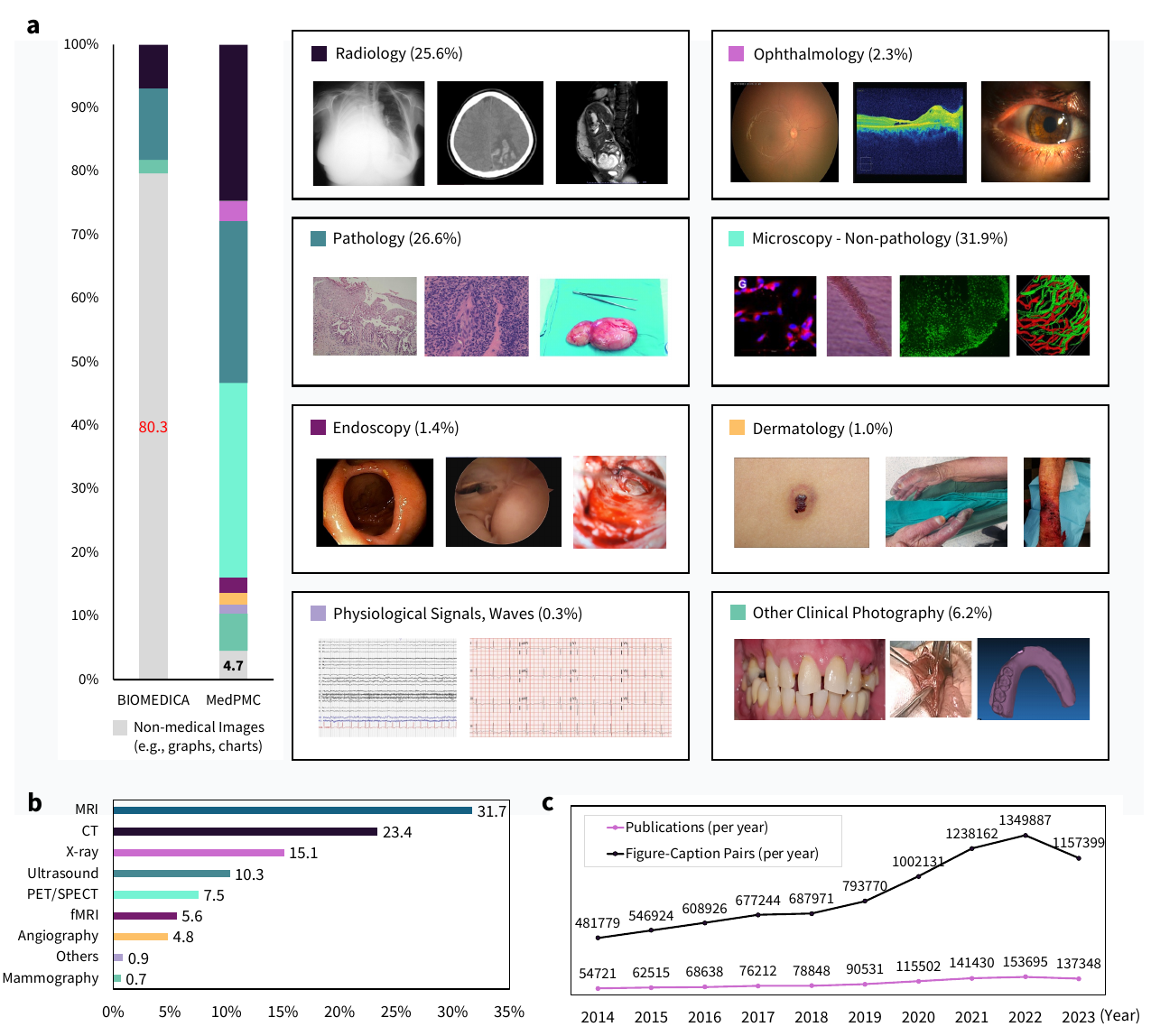}
\caption{
\textbf{Overview of MedPMC dataset composition, modality distribution, and temporal growth.}
\textbf{a}, Distribution of image categories in MedPMC compared to BIOMEDICA, highlighting a substantial reduction in non-medical images (e.g., graphs and charts) and increased coverage of clinically relevant modalities.
\textbf{b}, Distribution of radiology imaging modalities within the MedPMC dataset.
\textbf{c}, The number of publications and extracted figure-caption pairs per year using our framework, demonstrating its scalability and capacity for continual expansion.
}
\label{figure:data_analysis}
\end{figure}

%% file: figures/clip_results.tex
\begin{figure}[t!]
\centering
\includegraphics[width=0.99\linewidth]{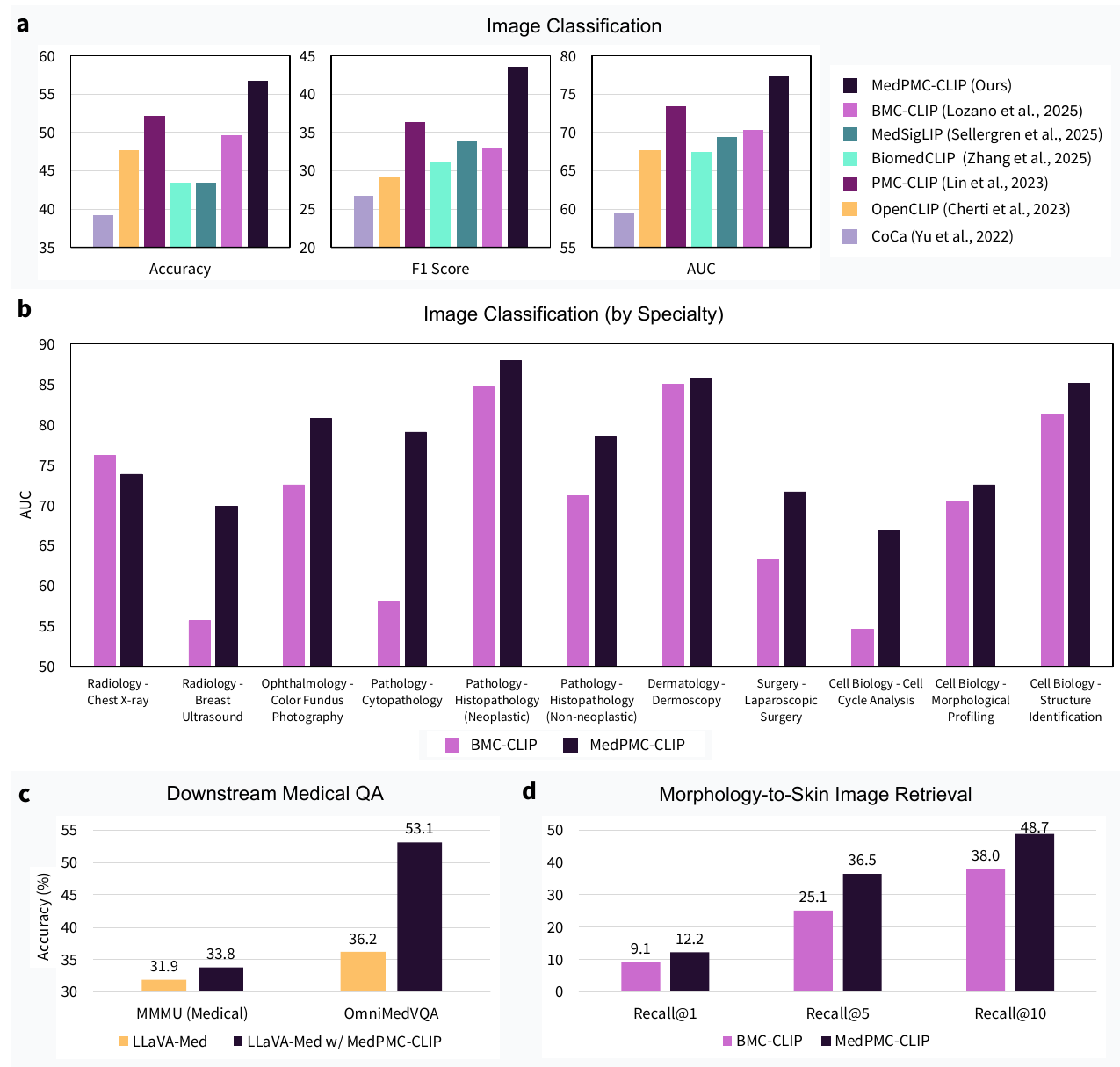}
\caption{
\textbf{Evaluation results.}
\textbf{a}, Image classification performance across 11 medical specialties, where scores are averaged over 26 benchmarks; MedPMC-CLIP consistently outperforms prior models across accuracy, F1 score, and AUC.
\textbf{b}, Comparison of MedPMC-CLIP and BMC-CLIP on image classification performance across medical specialties; MedPMC-CLIP outperformed BMC-CLIP in 10 of 11 specialties.
\textbf{c}, Performance comparison on multimodal medical QA benchmarks (MMMU and OmniMedVQA), showing substantial improvements when integrating MedPMC-CLIP into LLaVA-Med.
\textbf{d}, Morphology-to-skin image retrieval results, demonstrating improved representation quality with gains in Recall@1, Recall@5, and Recall@10 over BMC-CLIP.
Results with 95\% confidence intervals are provided in \textbf{Supplementary Information, Section 2}.
}
\label{figure:main_results}
\end{figure}

%% file: figures/visualization.tex
\begin{figure}[t!]
\centering
\includegraphics[width=\linewidth]{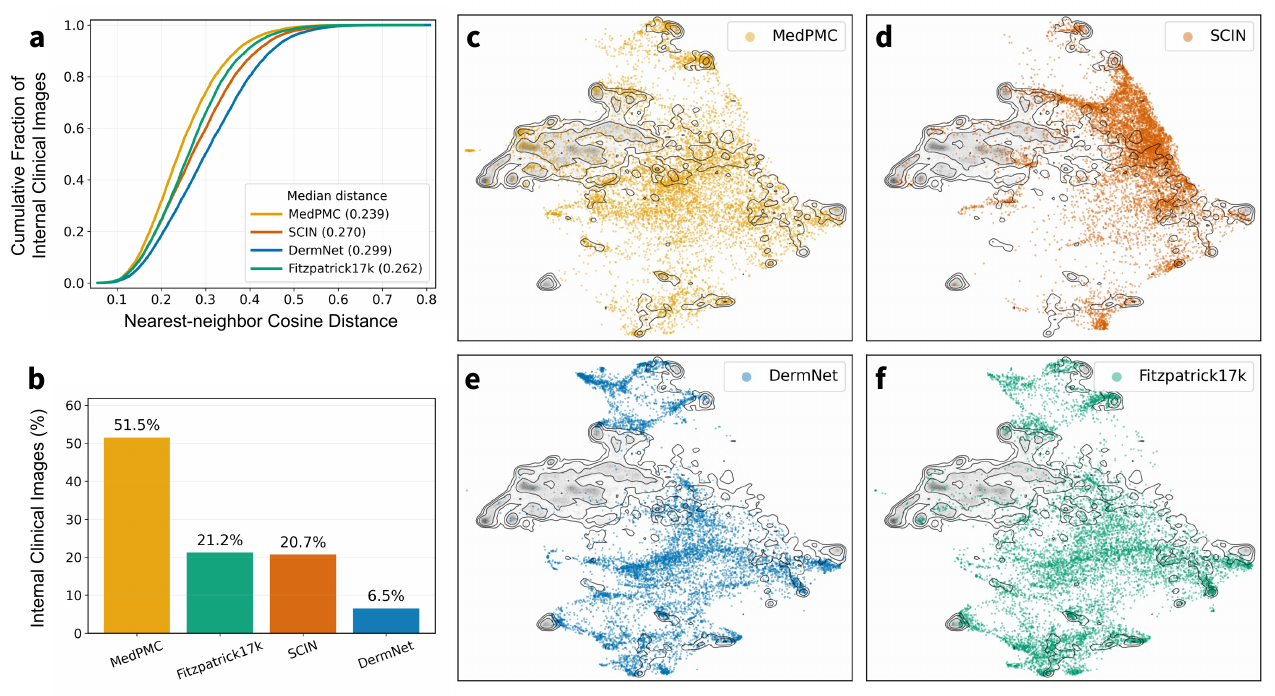}
\caption{
\textbf{Embedding-space comparison of external dermatology image sources with internal clinical dermatology images.}
\textbf{a}, Empirical cumulative distribution of nearest-neighbor cosine distances from internal clinical dermatology images to each external image source in the DINOv2 embedding space.
\textbf{b}, Fraction of internal clinical dermatology images whose nearest neighbor was found in each external source.
\textbf{c–f}, UMAP projections of image embeddings, showing the internal clinical dermatology image space with each external source overlaid separately.
Gray density and contour lines represent internal clinical dermatology images, and colored points represent images from each external source.
All panels were generated using DINOv2 image embeddings.
}
\label{figure:visualization}
\end{figure}

%% file: sections/discussion.tex
\section{Discussion}

We developed MedPMC, an automated and systematic framework for curating high-fidelity medical multimodal data from permissively licensed biomedical literature, together with MedPMC-CLIP, a vision-language foundation model trained on the resulting MedPMC corpus to demonstrate its utility.
The main finding is that biomedical literature, when transformed into medically relevant and panel-level aligned data, can support effective and reproducible development of medical multimodal models.
These data-level improvements translated into consistent model-level gains. 
Using the exact training protocol of BMC-CLIP and changing only the training corpus, MedPMC-CLIP improved average zero-shot AUC by 7.1 percentage points over BMC-CLIP across 26 public medical benchmarks spanning 11 specialties. 
Gains in AUC were observed in 10 of the 11 specialties, despite MedPMC-CLIP being trained on fewer than half as many image-text pairs. 
These benefits also extended to MLLMs: within the LLaVA-Med framework, replacing only the vision encoder with MedPMC-CLIP---while holding the remaining model and training pipeline fixed---improved performance by 1.9 percentage points on MMMU and 16.9 percentage points on OmniMedVQA.
Importantly, these improvements were not confined to public benchmarks. 
Using 10,524 dermatology patient records from a multi-site, hospital-based consultation cohort, MedPMC-CLIP improved performance in zero-shot retrieval.
Embedding-space analysis further showed that MedPMC dermatology images were more closely aligned with clinical dermatology images than several widely used public dermatology datasets.
These findings suggest that MedPMC can help address a central bottleneck in medical multimodal AI---the scarcity of large-scale, clinically meaningful multimodal data~\cite{zhang2024challenges,schouten2025navigating,acevedo2020dataset,huang2025systematic}. Beyond serving as a corpus, MedPMC provides a modular curation framework with component-level benchmarks that can be updated as new biomedical articles become available. Because it is derived from permissively licensed literature, the resulting resources can be shared more openly, supporting reproducibility, adaptation, and continued model development across both benchmark and clinically realistic settings.

\paragraph{Data curation as reusable infrastructure for medical AI}

A key implication of this study is that data curation should be treated as a core engineering problem in medical multimodal AI.
PMC offers a scalable public source of biomedical image-text data, and the resource continues to grow as new biomedical knowledge is published~\cite{pmc_intro}.
However, directly repurposing PMC figures introduces several failure modes, including non-medical visual content, compound multi-panel figures, weak subfigure-subcaption correspondence, and dataset staleness.
MedPMC addresses these failure modes through a modular five-stage pipeline.
The initial screening step identifies medically relevant figures before image download, reducing unnecessary data transfer and storage.
Multi-panel figure detection and separation convert compound figures into panel-level visual units.
Caption separation and subfigure-subcaption alignment improve image-text correspondence at the level at which supervision is most meaningful.
Medical figure classification further filters decomposed panels to remove remaining non-medical visual elements.
Because each stage is supported by component-level benchmarks, the pipeline can be evaluated, audited, and improved rather than treated as an opaque preprocessing procedure.
This infrastructure framing distinguishes MedPMC from a one-time dataset release.
Biomedical literature continues to expand, and static datasets can become progressively incomplete as new diseases, technologies, imaging modalities, and clinical terminology emerge.

To keep MedPMC current, we plan to release updated corpus versions every six months, aligned with the semiannual PMC baseline snapshots. After the initial June 2024 release, each update will process newly available permissively licensed articles not included in prior versions and will be documented by article cutoff date, license filters, pipeline component checkpoints, and processing configuration. Based on recent publication trends, these updates are expected to identify about one million additional medical image-text pairs per year. The modular design of MedPMC also allows individual pipeline components to be replaced or improved over time.

\paragraph{Relationship to existing PMC-derived resources}

MedPMC builds on a growing literature that has established biomedical articles as a valuable source of medical multimodal supervision.
Early resources such as ROCO~\cite{pelka2018radiology} and MedICaT~\cite{subramanian2020medicat} showed that figures and captions from biomedical literature could support medical vision-language learning, although these datasets were relatively small or focused on narrower domains.
More recent PMC-derived resources, including PMC-OA~\cite{lin2023pmc}, PMC-15M~\cite{zhang2025multimodal}, BIOMEDICA~\cite{lozano2025biomedica}, and Open-PMC-18M~\cite{baghbanzadeh2025open}, substantially expanded the scale of literature-derived image-text data, enabling medical vision-language models such as PubMedCLIP~\cite{eslami2023pubmedclip}, BiomedCLIP~\cite{zhang2025multimodal}, and BMC-CLIP~\cite{lozano2025biomedica}, as well as multimodal medical LLMs such as Med-Flamingo~\cite{moor2023med}, LLaVA-Med~\cite{li2023llava}, BiomedGPT~\cite{zhang2024generalist}, HealthGPT~\cite{lin2025healthgpt}, and MedGemma~\cite{sellergren2025medgemma}.
However, as detailed in \textbf{Extended Data Table~\ref{table:pmc_dataset_comparison}}, these large-scale data resources remain limited in medical relevance, compound-figure processing, and reproducible curation pipelines.

MedPMC extends this line of work by emphasizing fidelity, reproducibility, and updatability rather than scale alone.
First, many existing large-scale PMC-derived datasets include substantial non-medical visual content, such as graphs, charts, molecular illustrations, workflow diagrams, and system schematics.
In the manual analysis, ~80\% of sampled BIOMEDICA figures were non-medical, compared with 4.7\% for MedPMC.
Filtering strategies based on manually curated keywords can miss medically relevant figures outside predefined terminology, whereas image-only filtering requires large-scale image download before screening and may not capture clinical context.
MedPMC instead formulates medically relevant figure selection as a text classification problem using captions and inline reference text, which provides contextual information before image download and improves scalability.
This text-based classifier reached a recall of 93.1 and an F1 score of 93.2, substantially outperforming keyword-based filtering.
Second, existing resources differ substantially in how they handle compound figures.
Some exclude multi-panel figures, which discards informative medical images; others retain compound figures as whole figure-caption pairs, which can weaken image-text correspondence when panels depict different modalities, disease states, time points, or experimental conditions.
Panel-level approaches based on sequential caption splitting and heuristic or CLIP-based alignment can also introduce alignment errors, while approaches that run an MLLM separately for every subfigure~\cite{baghbanzadeh2025open} incur inference costs that scale with panel count.
MedPMC instead separates compound figures and jointly performs caption separation and subfigure-subcaption alignment with a supervised 4B-parameter MLLM that considers all panels together.
This approach reached 81.4 F1 and 85.3 ROUGE-L, outperforming rule- and CLIP-based matching used in prior pipelines as well as GPT-4, while remaining efficient at scale by avoiding per-subfigure inference.
Third, most prior PMC-derived resources are released as static snapshots.
Because their full curation pipelines, trained preprocessing models, and stage-specific evaluation resources are often not fully available, reproducing or extending the dataset construction process can be difficult.
MedPMC was designed to make the curation process itself reusable and to provide a reproducible framework for generating, evaluating, and updating medical multimodal supervision at scale.

\paragraph{Complementary strengths of MedPMC and clinical data}

Importantly, MedPMC is not intended to replace clinical datasets.
Literature-derived and clinical resources serve complementary roles in medical multimodal model development.
Clinical datasets remain closest to deployment because they capture real patient populations, local imaging protocols, clinical workflows, and case mix.
However, they are often concentrated within a limited number of institutions, specialties, and modalities, and their accompanying text is frequently limited to diagnostic labels or short reports rather than detailed descriptions of image findings.
Despite several pioneering efforts~\cite{irvin2019chexpert,huang2023inspect,ai2024ai,wang2024nci}, such datasets remain difficult to assemble and share at scale because of privacy, governance, licensing, and annotation constraints.

PMC-derived data may offer a distinct but complementary form of supervision. Compared with many clinical datasets, literature-derived resources can provide greater scale, richer image-associated context, including experimental settings and diagnostic descriptions, and broader coverage across diseases, specialties, modalities, and emerging concepts, including rare conditions and technologies that may be sparsely represented in existing clinical resources. Because they are derived from publicly available and permissively licensed literature, they can also be shared and reproduced more readily than many institution-specific clinical datasets. One practical development pathway is therefore to use MedPMC-like resources for broad pretraining and representation learning, followed by clinical adaptation and validation in the intended target setting. In this workflow, literature-derived data provide scale, coverage, and descriptive supervision, while clinical data provide local grounding and direct alignment with real-world practice. This complementary role is further supported by preliminary few-shot adaptation results in \textbf{Supplementary Information, Section 3}, which show a modest performance advantage over OpenCLIP and BMC-CLIP.

Our clinical dermatology evaluation demonstrates how literature-derived pretraining can support downstream clinical applications.
Although MedPMC is derived from biomedical publications rather than routine clinical workflows, MedPMC-CLIP transferred to morphology-guided retrieval on internal patient data.
The embedding-space analysis further suggested that MedPMC dermatology images overlap more broadly with internal clinical dermatology images than several widely used public dermatology datasets.
This finding does not imply that literature-derived images are equivalent to clinical data or sufficient for deployment without target-domain validation. Rather, it indicates that carefully curated biomedical literature can provide pretraining supervision that is relevant to downstream clinical tasks and can complement, rather than replace, institution-specific clinical datasets.

\paragraph{Potential applications}

First, the corpus can be used to train or initialize medical vision-language encoders, visual backbones for MLLMs, and specialty-specific foundation models~\cite{dai2024pa,qin2026volmo}.
Second, because each image is linked to article text, figure captions, subcaptions, and medical figure categories, MedPMC can be filtered into specialty-, modality-, or concept-specific subsets.
This may be useful when dedicated public datasets are limited, such as in rare diseases, surgical imaging, pathology subdomains, clinical skin photography, or specific imaging modalities~\cite{nievas2025open,wu2024gestaltmml}.
Third, the structure of compound figures provides a source of supervision that is difficult to obtain from standard single-image datasets.
Many biomedical figures show related images, such as disease progression, pre- and post-treatment comparisons, multi-view imaging, paired radiology-pathology findings, or contrasts between disease subtypes and controls.
These structures could support models that reason over image sets rather than isolated images~\cite{chen2025compound}.
Fourth, the aligned image-text pairs can be indexed as a searchable corpus of biomedical visual evidence.
Such an index could support similar-case search, visual example retrieval, and multimodal retrieval-augmented generation, in which MedPMC functions both as a training resource and as an external knowledge source for downstream multimodal systems~\cite{yuan2023ramm}.

\paragraph{Limitations and Future Work}

First, this study focuses on image-text data, whereas clinical practice routinely involves broader and more heterogeneous information streams, including structured tables, laboratory values, temporal clinical context, and longitudinal patient records. Some of these modalities, such as tables and structured content embedded in biomedical articles, could be incorporated into a similar literature-based curation framework. Extending MedPMC to capture these additional modalities is therefore an important direction for future work. Other data types, particularly longitudinal patient records and real-world clinical workflows, are rarely available in a consistent or shareable form in the public literature. Addressing these gaps will require complementary frameworks for curating and integrating clinical data under appropriate privacy, governance, and institutional constraints.

Second, because MedPMC is derived from biomedical articles, the corpus reflects both the strengths and limitations of the source literature. Biomedical figure captions vary in quality and may contain incomplete descriptions, inconsistent terminology, non-visual information, or diagnostic context that is not directly inferable from the image alone. In addition, biomedical publications introduce selection bias in what is visually reported. Published figures may preferentially show representative examples, diagnostically salient findings, unusual cases, positive or publishable results, or visually clear images, and may differ from routine clinical images in terms of case mix, acquisition setting, image quality, presentation style, and patient population~\cite{nissen2014clinical}. Prior work has shown that improving caption quality or restructuring text for multimodal learning can improve model performance~\cite{xu2024demystifying,chen2024towards}; these approaches are complementary to MedPMC because they depend on accurately aligned image-caption pairs as input. Future work could therefore combine MedPMC-style high-fidelity image-text alignment with caption refinement, structured report generation, and region- or concept-level supervision.

Third, although we systematically evaluated MedPMC across different model types, including contrastive vision-language encoders and MLLMs, and across multiple settings, including public benchmark evaluations, downstream tasks, and clinically realistic applications, our experiments do not exhaustively cover all possible model architectures or applications. This reflects the role of MedPMC as reusable data infrastructure rather than as a single task-specific model. Given the rapid development of medical foundation models, it is not feasible for one study to evaluate every emerging architecture or clinical use case~\cite{bedi2025testing,bedi2026holistic,chen2025benchmarking}. In addition, potential overlap between large-scale pretraining corpora and public benchmarks is an increasingly recognized challenge for foundation-model and LLM evaluation~\cite{sheikhi2026beyond,li2026memorization,chang2024survey}. We carefully selected public benchmarks that were largely derived from institutional clinical collections, established medical imaging repositories, or challenge datasets rather than directly from PMC articles, and we further included an internal YNHHS clinical cohort that was independent of the PMC-derived pretraining corpus. Nevertheless, exact overlap cannot be fully excluded. We therefore interpret the public benchmark results together with architecture-matched comparisons, downstream QA evaluation, and independent clinical transfer experiments.

Fourth, although our experiments show that models trained with MedPMC transfer to clinically motivated tasks, including morphology-guided dermatology image retrieval using internal patient data, these evaluations do not establish deployment readiness. MedPMC should be viewed as a source of broad pretraining supervision that complements, rather than replaces, target-domain clinical data. Additional clinical evaluations of model adaptation, such as further fine-tuning, and clinician-in-the-loop workflows will also be important for quantifying how public medical resources and models derived from them adapt to real-world clinical settings and support clinical decision-making~\cite{liu2019comparison}.
By releasing the corpus, curation pipeline, component-level benchmarks, pretrained models, and source-license metadata, we aim to support continued external evaluation, model comparison, pipeline refinement, and clinically grounded adaptation by the research community.

%% file: sections/methods.tex
\section{Methods}

\subsection{MedPMC Pipeline Implementation}

We trained specialized models for each stage of the MedPMC curation pipeline.
For each task, we leveraged existing datasets when they were well aligned with the target prediction problem.
However, available datasets were often limited in scale, restricted to specific domains, or not fully representative of PMC figure layouts and captioning styles.
To improve robustness in the target curation setting, we additionally constructed new datasets.
Dataset statistics for each pipeline component are provided in \textbf{Extended Data Table~\ref{table:pipeline_train_test_data}}.
The detailed prompts used for synthetic data generation are provided in \textbf{Supplementary Information, Section 4}.
Importantly, all datasets, trained models, and code for running and evaluating each pipeline component are publicly released, supporting reproducibility, reuse, benchmarking, and transparent inspection of implementation details.

\paragraph{Initial screening}
The initial screening model was designed to identify candidate medical figures before image download. For each figure, we concatenated the figure caption with the inline reference text from the article body and formulated the task as binary classification of medical relevance. To construct the training data, we randomly sampled 39,360 figures from articles indexed by Open-i (\url{https://openi.nlm.nih.gov}), an open-access biomedical image search engine, using the ``Exclude Graphics'' option to reduce the retrieval of plots, charts, and other non-image graphics. We then generated synthetic medical-relevance labels for the sampled figures using GPT-4T (\texttt{gpt-4-turbo-2024-04-09}).
Following our definition of medical figures, GPT-4T assigned labels after jointly considering both the image and its associated text. We initialized the classifier with PubMedBERT~\cite{gu2021domain}, which was pretrained on PubMed abstracts and full-text articles (\url{https://huggingface.co/microsoft/BiomedNLP-BiomedBERT-base-uncased-abstract-fulltext}).
The model was fine-tuned with a batch size of 64 and a learning rate of 3e-5.

\paragraph{Multi-panel figure detection}
The multi-panel figure detection model was designed to distinguish single-panel from multi-panel figures before subfigure separation. We formulated this step as binary image classification and trained the model using 16,775 samples from the compound figure detection subtask of the ImageCLEFmed challenge, together with 39,360 additional synthetically labeled examples. The synthetically labeled examples were drawn from the same Open-i sample used for initial screening, but were re-annotated for panel structure rather than medical relevance. GPT-4 was provided with each figure and its caption and assigned a binary label indicating whether the figure contained a single panel or multiple panels.
We adopted a Vision Transformer (\url{https://huggingface.co/timm/vit_base_patch16_rope_reg1_gap_256.sbb_in1k}) as the backbone model~\cite{dosovitskiy2021image} and fine-tuned the model using the PyTorch Image Models repository (timm)~\cite{rw2019timm} for 100 epochs with a batch size of 64. 
All other hyperparameters followed the default settings of the training framework, with the learning rate automatically scaled according to the effective global batch size.

\paragraph{Multi-panel figure separation}
The multi-panel figure separation model took figures classified as multi-panel as input and predicted bounding boxes corresponding to individual subfigures.
For training, we used the dataset provided for the figure separation task in the ImageCLEFmed challenge~\cite{GSB2016}.
We did not incorporate additional synthetic annotations at this stage because they did not further improve validation performance.
We used YOLOv10 as the backbone model~\cite{wang2024yolov10}.
The model was initialized from pretrained weights and fine-tuned using the default training configuration in Ultralytics v8.2.73 (\url{https://github.com/ultralytics/ultralytics}). 
Training was performed for 100 epochs with a batch size of 16 and an input resolution of 640 × 640; automatic optimizer selection chose AdamW with an initial learning rate of 2e-3~\cite{loshchilov2017decoupled}.

\paragraph{Caption separation and alignment}
The caption separation and alignment model took as input the original multi-panel figure, its full caption, and the separated subfigures, and generated the corresponding subcaptions. For training, we used 1,361 manually annotated samples from the MedICaT dataset~\cite{subramanian2020medicat}. 
To improve scale and domain diversity beyond the predominantly radiology-focused MedICaT dataset, we generated an additional 20,601 synthetic annotation examples. 
Specifically, we applied the trained multi-panel detection model to the Open-i figures used in the previous stage, selected those predicted as multi-panel, and separated them into subfigures using the trained YOLOv10 model.
GPT-4 was then provided with the original figure, the separated subfigures, and the full figure caption to generate aligned subcaptions for each subfigure.
We adopted InternVL-2.5-4B-MPO as the backbone model~\cite{chen2024expanding}.
The model was then trained using the official implementation of InternVL (\url{https://github.com/OpenGVLab/InternVL}) with the default training configuration, using three training epochs, a total batch size of 128, a learning rate of 4e-5, and BF16 mixed-precision training.

\paragraph{Medical figure classification}
The medical figure classification model performed binary image classification to determine whether an input image or subfigure was medically relevant.
We constructed a training set of 26,453 samples by combining data from ImageCLEF, MedICaT, and DocFigure~\cite{jobin2019docfigure}.
A Vision Transformer was used as the backbone model.
We trained the model using the timm library with the same setup as the multi-panel figure detection model, except that training was performed for 300 epochs.

\paragraph{Validation and model selection}

For each pipeline stage, we evaluated multiple backbone models, input configurations, and other task-specific design choices, and selected the final component based on validation-set performance. The validation sets combined existing benchmark data with PMC-derived synthetic annotations generated using the same procedures as the corresponding training data. To prevent overlap between training and validation samples originating from the same source article, synthetic annotations were partitioned at the article level. Validation-set composition and sample counts are provided in \textbf{Extended Data Table~\ref{table:pipeline_train_test_data}}.
Initial screening, multi-panel figure detection, and medical figure classification were evaluated using precision, recall, and F1 score. Following prior work~\cite{meng2024yolo}, multi-panel figure separation was evaluated using accuracy and mean average precision (mAP): accuracy measured agreement between the predicted and reference panel structure, whereas mAP quantified the localization of predicted subfigure bounding boxes relative to the ground-truth annotations. Following MedICaT, the caption separation and alignment task was evaluated using F1 score and ROUGE-L, both computed between the predicted and reference subcaptions.
The selected models substantially outperformed the comparison methods across the pipeline stages. Detailed descriptions of the baseline models and comparative analyses are provided in \textbf{Supplementary Information, Section 1}.

\subsection{Benchmark Annotation}

To enable systematic evaluation of the five components of the MedPMC pipeline, we constructed dedicated test sets using existing benchmark data and newly annotated examples from PMC. Whereas the validation sets were used for model selection, hyperparameter tuning, and other development decisions, test sets were kept fully held out and used only for final performance evaluation. Newly annotated PMC samples were drawn from articles indexed by Open-i, following the same initial sampling strategy used for training-data construction. However, all articles represented in the test sets were excluded from both the training and validation sets to prevent article-level overlap.

The manual annotation covered the first four stages: (1) initial screening, (2) multi-panel figure detection, (3) multi-panel figure separation, and (4) caption separation and alignment. 
For the medical figure classification task, manual annotation was bypassed because a substantial dataset of approximately 19,000 samples had already been secured from public resources.
New annotations were performed by 13 annotators (S.S.A., Y.W., Y.K., T.G., Z. Cao., M.D., Y.F., Y.H., R.S., G.Y., K.W.J., Y.L., and Y.T.). Before annotation, all annotators completed a training session covering the annotation guidelines and representative examples. Each sample was independently annotated by two annotators, and disagreements were adjudicated by a senior annotator (H.K.). After each annotation batch, the adjudicated decisions and their rationales were shared with the annotators to promote consistency across subsequent batches.

For the initial screening, four annotators (S.S.A., Z. Cao., Y.F., and Y.H.) reviewed image-caption pairs to annotate each sample as either medical or non-medical. Under our operational definition of medical figures, non-human-centered subjects, such as animal models, were strictly classified as non-medical. For this task, a total of 600 samples---comprising a mixture of single-panel and multi-panel figures to faithfully reflect real-world data distributions---were partitioned into batches of 100. 
Each annotator evaluated three separate batches, culminating in 300 annotations per annotator.
For multi-panel figure detection, two annotators (R.S. and G.Y.) labeled a total of 500 figures as either single-panel or multi-panel. This dataset was partitioned into five batches of 100 figures each.
The multi-panel figure separation and caption separation and alignment stages were performed concurrently by the same cohort of annotators within a unified interface. For this joint task, seven annotators (Y.W., Y.K., T.G., M.D., K.W.J., Y.L., and Y.T.) processed 455 compound figures across six batches. For each figure, annotators delineated the bounding boxes of individual subfigures and subsequently extracted and aligned the corresponding subcaptions. If a figure was determined not to be multi-panel, this workflow was skipped.

\subsection{Dataset Composition Analysis}

MedPMC consists of multi-panel figures, single-panel figures, and subfigures extracted from multi-panel figures.
For dataset composition analysis, we excluded multi-panel figures because assigning a single category to a compound figure is inherently ambiguous.
We randomly sampled 1,232 single-panel figures and 1,674 subfigures from MedPMC, totaling 2,906 samples.
Each group was evaluated separately, and the final distribution was obtained by weighting the results according to the proportions of single-panel figures and subfigures in the full dataset.
For BIOMEDICA, we similarly excluded multi-panel figures and restricted the analysis to single-panel images.
To do so, we applied our compound figure detection model to identify and filter out multi-panel images, resulting in 9.3 million images (38.7\%) retained for analysis.
From this subset, we randomly sampled 432 images for manual evaluation.

A total of five annotators (P.X., S.S.A., R.J., Y.D., and J.M.) participated in the study, including two hospital residents and one final-year medical student.
Each annotator was presented with both the image and its corresponding caption through a dedicated annotation interface.
Annotators were asked to assign each sample to one of the predefined categories.
If a sample did not fit any predefined category, annotators could select an ``Other'' option and provide a brief specification.
The three medical experts additionally reviewed and resolved ambiguous cases that were initially marked as uncertain.
The category schema was designed in collaboration with the medical experts, and the definition of non-medical images was informed by prior work~\cite{GSB2016}.

\subsection{Model Training and Downstream Evaluations}

\paragraph{Baseline models}
We compared MedPMC-CLIP against a range of baseline vision-language models for image classification and retrieval. These include general-domain foundation models such as (1) {CoCa}~\cite{yu2022coca}, a contrastive-captioning model trained on web-scale datasets, and (2) {OpenCLIP}~\cite{cherti2023reproducible}, a standard CLIP architecture trained on LAION-2B. 
Within the medical domain, we evaluated (3) PMC-CLIP~\cite{lin2023pmc}, specifically the version initialized from the OpenAI CLIP ViT-L/14 checkpoint and subsequently fine-tuned on PMC-OA and ROCO data, (4) {BiomedCLIP}~\cite{zhang2025multimodal} pretrained on the PMC-15M corpus, and (5) {MedSigLIP}~\cite{sellergren2025medgemma}, a sample-efficient medical encoder optimized via sigmoid pairwise loss. 
Finally, we compared our model against (6) BMC-CLIP~\cite{lozano2025biomedica}, which was trained on the BIOMEDICA-24M dataset; this model serves as the most direct baseline for evaluating MedPMC-CLIP, as we maintained the same model architecture and training pipeline configuration while isolating the pretraining dataset as the sole variable.
For medical visual QA, we used LLaVA-Med~\cite{li2023llava}, a biomedical adaptation of LLaVA (v1.5)~\cite{liu2023visual} that connects a CLIP vision encoder to a language model through a linear projection layer and is trained through a two-stage curriculum of visual alignment and instruction tuning. 

\paragraph{Training details}
For MedPMC-CLIP, we followed the experimental setup of BIOMEDICA using the official code repository (\url{https://github.com/minwoosun/biomedica-etl}).
Specifically, we initialized the model from OpenCLIP (ViT-L/14; checkpoint: \texttt{commonpool\_xl\_clip\_s13b\_b90k} and adopted the same configuration and hyperparameters (a batch size of 8,192 and a learning rate of 1e-6), while replacing the training data with MedPMC (11 million image-text pairs)
For MLLM evaluation, we adopted the same training data and hyperparameters as LLaVA-Med~\cite{li2023llava}, following the official implementation (\url{https://github.com/microsoft/LLaVA-Med/tree/main}).
Stage~1 (biomedical concept feature alignment) was performed on 467K samples using a batch size of 128 and a learning rate of 2e-3.
Stage~2 (end-to-end instruction tuning) was then conducted on 56K samples for 3 epochs using a batch size of 128 and a learning rate of 2e-5.
We used the LLaVA training codebase (\url{https://github.com/haotian-liu/llava}).

\paragraph{26 image classification benchmarks}
We used a comprehensive suite of medical benchmarks covering a diverse range of clinical modalities and specialties, including radiology, ophthalmology, pathology, dermatology, surgery, and microscopy. 
Detailed specifications regarding benchmark specialties, core tasks, and sample sizes are provided in \textbf{Extended Data Table~\ref{table:list_of_benchmarks}}. 
For each task, prompt templates were adopted directly from those supported by Micro-Bench~\cite{lozano2024micro}; for domains not covered by existing frameworks, we custom-designed the evaluation prompts, all of which are publicly verifiable via our released evaluation code (\url{https://github.com/Yale-BIDS-Chen-Lab/MedPMC}). 

Performance was evaluated using accuracy, F1 score, and area under the receiver operating characteristic curve (AUC). 
For binary and multiclass tasks, accuracy was calculated using the top-scoring class, F1 score was macro-averaged across classes, and AUC was calculated by pooling the one-hot-encoded labels and corresponding scores across all sample-class pairs. 
For multi-label tasks, predictions were obtained using a threshold of 0.5, and accuracy was calculated across all sample-label pairs. 
Positive-class F1 scores and AUCs were calculated independently for each label and then macro-averaged across labels; labels without both positive and negative samples were excluded from AUC calculation. 

\paragraph{Downstream medical QA}
We evaluated MLLMs on MMMU~\cite{yue2024mmmu} and OmniMedVQA~\cite{hu2024omnimedvqa} in a zero-shot setting. These benchmarks were selected because they are highly popular and widely recognized standards within the community, providing an exceptionally broad coverage of the medical domain rather than focusing on a single, narrow specialty.
MMMU spans a diverse set of 30 domains; in our experiments, we focused on the clinical medicine subset and included only single-image questions, consisting of 314 samples.
OmniMedVQA originally comprises more than 120,000 samples from 73 medical datasets, subdivided into 42 open-access and 31 restricted-access repositories. 
In this study, we used only the open-access datasets and excluded four COVID-19-related datasets curated from the literature to minimize the risk of data contamination (see ``Contamination-aware evaluation design''), resulting in 38 open-access datasets.
To ensure balanced and computationally efficient evaluation across these various source datasets, we randomly sampled a maximum of 100 examples per dataset.
We then removed DRIMDB samples that assessed about image quality (good, bad, or outlier) but were incorrectly categorized as lesion grading, resulting in a final evaluation set of 3,593 samples.
For all inference generations, we utilized greedy decoding with the temperature fixed at 0.0 to ensure deterministic results.
Models were evaluated using only the image and the original question, with no prompting engineering.

\paragraph{Downstream clinical application: morphology-to-skin-image retrieval}

Retrieving similar medical images has long been studied as a form of decision support and education in dermatology and broader medical imaging contexts~\cite{sadeghi2020using,gassner2023saliency,muller2004review,long2009content}.
In this study, textual morphology descriptions were used as queries to retrieve relevant clinical dermatology images.
The study was approved by the Yale Institutional Review Board (IRB) under protocol number 2000041054.
We constructed the evaluation dataset from 10,524 real-world dermatology patient images from a multi-site, hospital-based dermatology consultation cohort within the Yale New Haven Health System (YNHHS).
The dataset included multiple images from some patients, although each image was treated as an individual retrieval sample. All models were evaluated using the same set of images.
For each sample, a clinical image was originally paired with a raw morphology description extracted from the corresponding clinical note.
Because multiple clinical images can share the same morphological characteristics, using only the original, unnormalized image-description pairs as ground truth would underestimate clinically relevant matches.
To address this and standardize the textual descriptions, the raw morphology descriptions were automatically mapped to standardized morphology concepts based on the SkinCon taxonomy~\cite{daneshjou2022skincon} (e.g., papule, plaque, macule, and erythema) using an open-sourced large language model, gpt-oss-120b~\cite{agarwal2025gpt}.
To ensure data quality and evaluate the reliability of this automated normalization, a hospital resident in dermatology (R.J.) manually reviewed and verified a sampled subset of the generated mappings. 
The detailed prompt templates for this LLM-based parsing are provided in \textbf{Supplementary Information, Section 4}.
This standardization mapping enabled images sharing the same set of morphology concepts to be grouped as clinically related.
For retrieval evaluation, we adopted an established set-based matching criterion~\cite{liu2016multi,kuster2008evaluating}.
A retrieved image was considered correct if its associated morphology concept set contained the ground-truth concept set of the query image.
This formulation allowed multiple retrieved images to be treated as correct, reflecting cases in which different images shared the same core morphological features even when additional attributes differed.

\paragraph{Embedding-space comparison with clinical dermatology images}

We further examined whether dermatology images curated from MedPMC occupy a visual space similar to that of real-world clinical dermatology images.
To identify dermatology images within MedPMC, we used the dermatology-related categories and text prompts described in \textbf{Extended Data Table~\ref{table:data_analysis}}.
All MedPMC images were scored using MedPMC-CLIP based on image-text similarity, and images classified as skin photographs were used to construct the MedPMC dermatology subset.
We compared this subset against three widely used public dermatology datasets: Fitzpatrick17k~\cite{groh2021evaluating}, a collection of community-contributed dermatology photographs annotated with skin type and disease labels; SCIN~\cite{ward2024creating}, a dataset of smartphone-acquired skin images collected from diverse participants in real-world settings; and DermNet~\cite{dermnet}, a large collection of educational dermatology images curated from the DermNet resource.
To enable balanced comparisons across sources, 10,000 images were randomly sampled from each external dataset.
YNHHS clinical dermatology images served as the reference distribution.

All images were embedded using DINOv2 (ViT-B/14)~\cite{oquab2023dinov2}.
To quantify similarity between each external source and the YNHHS clinical image distribution, we computed the cosine distance from each YNHHS clinical image to its nearest neighbor within each external dataset.
We then summarized the resulting nearest-neighbor distance distributions and compared median distances across datasets.
In addition, each YNHHS clinical image was assigned to the external source containing its overall nearest neighbor across all external datasets, and the fraction of images assigned to each source was calculated.
For visualization, DINOv2 embeddings were projected into two dimensions using UMAP~\cite{mcinnes2018umap}.
The UMAP projection was fit on the combined image set and used to visualize the distribution of YNHHS clinical dermatology images together with images from each external source.

\paragraph{Statistical analysis}
We estimated 95\% CIs for AUC using nonparametric bootstrap resampling with 10,000 replicates. 
In each replicate, samples were drawn with replacement from the corresponding test set, and AUC was recalculated using the same evaluation procedure as for the point estimate. 
Binary and multiclass datasets were resampled within each class to preserve the original class distribution, whereas multilabel datasets were resampled at the sample level while retaining all labels associated with each selected sample. 
To compare MedPMC-CLIP with BMC-CLIP, identical bootstrap samples were applied to both models within each benchmark. For each bootstrap replicate, benchmark-level differences in accuracy, macro F1 score, and AUC were calculated, averaged within each specialty, and then macro-averaged across the 11 specialties.
For the medical QA experiments, individual model accuracies were reported with Wilson 95\% CIs. 
Between-model accuracy differences were evaluated using paired bootstrap resampling with 10,000 replicates, with identical question samples applied to baseline LLaVA-Med and LLaVA-Med with the MedPMC-CLIP encoder. 
Paired differences were calculated for the overall MMMU and OmniMedVQA results and for each reported OmniMedVQA query category.
For the morphology-guided retrieval experiment, differences in Recall@1, Recall@5, and Recall@10 were evaluated using query-level paired bootstrap resampling with 10,000 replicates.
All 95\% CIs were defined by the 2.5th and 97.5th percentiles of the corresponding bootstrap distributions. 
The full results are provided in \textbf{Supplementary Information, Section 2}. We also release the evaluation code in our official GitHub repository to support reproducibility.

\paragraph{Contamination-aware evaluation design}

Potential data contamination between large-scale pretraining corpora and public evaluation benchmarks is an increasingly recognized challenge, especially in the evaluation of foundation models and large language models~\cite{sheikhi2026beyond,li2026memorization,chang2024survey}. We therefore considered dataset provenance when designing the evaluation strategy. Most evaluation datasets used across the 26 benchmark tasks and downstream medical QA experiments were derived from institutional clinical collections, established medical imaging repositories, or challenge tasks rather than directly from PMC articles, which reduces the likelihood of systematic overlap with the MedPMC pretraining corpus. 
For OmniMedVQA, we further excluded four open-access subsets (``{Covid CT},'' ``CoronaHack,'' ``Covid19 heywhale,'' and ``COVIDx CXR-4'') that were curated from the literature, as such datasets are more likely to contain images from preprints or publications that may overlap with PMC-indexed articles.
However, provenance alone cannot guarantee complete image-level independence. Public and web-curated benchmarks may aggregate images from heterogeneous sources, and individual images can be reproduced across preprints, publications, repositories, and benchmark datasets. 

For this reason, we evaluated MedPMC-derived models across multiple complementary settings rather than relying solely on public benchmark performance. These included controlled architecture-matched comparisons on public benchmarks, downstream medical QA, and a clinical transfer evaluation using 10,524 dermatology patient images from the Yale New Haven Health System. The YNHHS cohort was not part of the PMC-derived pretraining corpus and therefore provides an independent assessment of transfer to clinically realistic patient data. The consistent gains observed across these evaluation settings suggest that MedPMC improves medical visual representations beyond potential benchmark-specific overlap.

%% file: sections/extended_data.tex
\input{tables/pmc_datset_comparison}
\input{tables/pipeline_train_test_data}
\input{tables/data_analysis_details}

\input{tables/list_of_benchmarks}

%% file: tables/pmc_datset_comparison.tex
\begin{table}[t]
\centering
\footnotesize
\begin{tabular}{lllllllll}
\toprule
\textbf{Dataset}  & \textbf{Release} & \textbf{Domain} & \textbf{\# Pairs} & \textbf{Filtering} & \begin{tabular}[c]{@{}l@{}}\textbf{Multi-panel}\\\textbf{Handling}\end{tabular} & \begin{tabular}[c]{@{}l@{}}\textbf{Dataset}\\\textbf{Update}\end{tabular} & \begin{tabular}[c]{@{}l@{}}\textbf{Pipeline}\\\textbf{Release}\end{tabular} & \begin{tabular}[c]{@{}l@{}}\textbf{Benchmark}\\\textbf{Release}\end{tabular} \\
\midrule
ROCO~\cite{pelka2018radiology} & 2018 & Radiology & 0.08M & $\triangle$ & $\times$ & Static & $\times$ & $\times$ \\
MedICaT~\cite{subramanian2020medicat} & 2020 & Medicine & 0.22M & $\triangle$ & $\triangle$ & Static & $\times$ & $\times$ \\
PMC-OA~\cite{lin2023pmc} & 2023 & Medicine & 1.65M & $\triangle$ & $\triangle$ & Static & $\times$ & $\times$ \\
PMC-15M~\cite{zhang2025multimodal} & 2023 & PMC & 15M & $\times$ & $\times$ & Static & $\times$ & $\times$ \\
BIOMEDICA~\cite{lozano2025biomedica} & 2025 & PMC & 24M & $\times$ & $\times$ & Static & $\times$ & $\times$ \\
Open-PMC-18M~\cite{baghbanzadeh2025open} & 2025 & Medicine & 18M & $\triangle$ & $\triangle$ & Static & $\times$ & $\times$ \\
\midrule
MedPMC & 2026 & Medicine & 11M & \ding{51} & \ding{51} & Continuous & \ding{51} & \ding{51} \\
\bottomrule
\end{tabular}
\caption{
{Comparison of PMC-based image-text datasets.}
We summarize prior datasets with respect to their initial release year, domain, scale, use of filtering, support for multi-panel figure and caption handling, dataset updating, release of the curation pipeline, and availability of human-annotated benchmarks for evaluating the curation process.
$\times$: not supported or excluded.
$\triangle$: supported with limited or coarse processing.
\ding{51}: supported using recent models and advanced methodologies.
Further details on the curation approaches adopted in existing datasets are provided in \textbf{Supplementary Information, Section 1}.
}
\label{table:pmc_dataset_comparison}
\end{table}

%% file: tables/pipeline_train_test_data.tex
\begin{table}[t]
\centering
\footnotesize
\begin{tabular}{llll}
\toprule
\multirow{1}{*}{\textbf{Stage}} & \multicolumn{1}{l}{\textbf{Training}} & \multicolumn{1}{l}{\textbf{Validation}} & \multicolumn{1}{l}{\textbf{Test}} \\
\midrule
Initial Screening & 39,360 (S) & 9,781 (S) & 
600 (L) \\
\midrule
\begin{tabular}[c]{@{}l@{}}Multi-panel Figure\\Detection\end{tabular} & 
39,360 (S),
16,775 (I) & 9,781 (S), 4,210 (I) & 
3,456 (I),
500 (L)  \\
\midrule
\begin{tabular}[c]{@{}l@{}}Multi-panel Figure\\Separation\end{tabular} & 6,103 (I) & 678 (I) & 
1,615 (I), 455 (L) \\
\midrule
\begin{tabular}[c]{@{}l@{}}Caption Separation\\\& Alignment\end{tabular} & 
20,601 (S), 1,664 (M) & 
418 (M) & 
455 (L) \\
\midrule
\begin{tabular}[c]{@{}l@{}}Medical Figure\\Classification\end{tabular} & \begin{tabular}[c]{@{}l@{}}5,477 (I), 5,136 (M),\\ 15,840 (D)\end{tabular} & \begin{tabular}[c]{@{}l@{}}1,299 (I), 1,214 (M),\\3,956 (D)\end{tabular} & \begin{tabular}[c]{@{}l@{}}4,166 (I), 1,659 (M), \\3,172 (D)\end{tabular} \\
\bottomrule
\end{tabular}
\caption{
Dataset statistics across the training, validation, and testing splits for each pipeline stage. 
Data types vary by source: Synthetic data (S), ImageCLEF (I), MedICaT (M), DocFigure (D), and newly annotated samples (L).
}
\label{table:pipeline_train_test_data}
\end{table}

%% file: tables/data_analysis_details.tex
\begin{table}[t]
\centering
\footnotesize
\begin{tabular}{llrr}
\toprule
\multirow{4}{*}{\textbf{Category}} & \multirow{4}{*}{\textbf{Sub-category}} & \multicolumn{2}{c}{\textbf{Image Distribution}} \\
\cmidrule(lr){3-4}
&  & \textbf{MedPMC} & \textbf{BIOMEDICA} \\
\midrule
Radiology & X-ray & \textbf{3.854}\% & 2.546\%  \\
& CT & \textbf{5.976}\%  & 1.157\% \\
& MRI & \textbf{8.108}\%  &  1.157\% \\
& fMRI & 1.434\% & 0.0\%  \\
& Mammography  & 0.171\% & 0.0\% \\
& PET/SPECT & 1.926\% &  0.0\% \\
& Ultrasound  & 2.631\% & 1.620\%  \\
& Angiography & 1.237\% & 0.463\%  \\
& Others & 0.220\% & 0.0\% \\
{Ophthalmology} & OCT & 0.883\% & 0.0\% \\
& Fundus photograph & 0.839\% & 0.0\% \\
& Slit-lamp photograph & 0.437\%  & 0.0\%  \\
& Others & 0.116\% & 0.0\% \\
{Pathology} & Histopathology/Cytology/Immunohistochemistry & \textbf{24.812}\% & \textbf{11.111}\%  \\
& Pathology gross photography & 1.795\% & 0.231\% \\
& Others & 0.0\%& 0.0\% \\
{Microscopy (non-pathology)} & - & \textbf{31.927}\% & 0.231\% \\
{Endoscopy} & - & 1.408\% & 0.0\% \\
{Dermatology} & Dermoscopy & 0.068\% & 0.0\% \\
& Skin photography  & 0.922\% & 0.0\% \\
& Total body photography & 0.0\% & 0.0\% \\
& Others & 0.055\% & 0.0\% \\
{Physiological Signals \& Waves} & Electroencephalography (EEG) & 0.108\% & 0.0\%\\
& Electrocardiography (ECG) & 0.193\% & 0.0\%  \\
& Electromyography (EMG) & 0.007\%  & 0.0\% \\
& Others & 0.020\% & 0.0\%\\
{Other clinical photography} & Clinical photograph & \textbf{5.564\%} & 1.157\% \\
& 3D reconstruction & 0.599\% & 0.0\% \\
\midrule
Non-medical Images & Table or form & 0.020\% & 6.019\%  \\
& Program listing & 0.055\% &  0.231\% \\
& Statistical figure, graph, or chart & 3.183\% & \textbf{41.667}\% \\
& Screenshot & 0.013\% & 0.0\%  \\
& Flowchart &  0.033\% &  \textbf{9.722}\%  \\
& System overview & 0.147\% & 4.630\%  \\
& Gene sequence & 0.0\% & 0.231\% \\
& Chromatography, gel & 0.013\%  & 0.0\% \\
& Chemical structure & 0.0\% & \textbf{9.028}\%  \\
& Mathematics or formula & 0.007\% & 0.231\% \\
& Hand-drawn sketch & 0.149\% & 0.0\% \\
& Natural image & 0.273\% & 0.0\% \\
& Others & 0.794\% & \textbf{8.565}\% \\
\bottomrule
\end{tabular}
\caption{
Manual evaluation of randomly sampled images from our MedPMC dataset and BIOMEDICA for image distribution analysis.
The top five most frequent categories are highlighted in bold.
}
\label{table:data_analysis}
\end{table}

%% file: tables/list_of_benchmarks.tex
\begin{table}[t]
\centering
\footnotesize
\begin{tabular}{llll}
\toprule
\textbf{Specialty / Modality} & \textbf{Task} & \textbf{Dataset} & \begin{tabular}[c]{@{}l@{}} \textbf{\# Samples}\\ \textbf{(Classes)}\end{tabular} \\
\midrule
{Radiology} \\
\quad Chest X-ray & Pneumonia detection & RSNA2018~\cite{shih2019augmenting} & 26,684 (2) \\
& Chest X-ray finding classification & ChestMNIST~\cite{medmnistv1} & 22,433 (14) \\
\cmidrule(lr){2-4}
\quad Breast ultrasound & Breast malignancy classification & BreastMNIST~\cite{medmnistv1} & 156 (2) \\
\midrule
Ophthalmology\\
\quad Color fundus photography & Diabetic retinopathy grading & DeepDRiD~\cite{liu2022deepdrid} & 400 (5) \\
\midrule
Pathology \\
\quad Cytopathology & White blood cell type classification & Acevedo et al.~\cite{acevedo2020dataset} & 1,600 (6) \\
& White blood cell type classification & Jung et al.~\cite{jung2022wbc} & 1,000 (6) \\
& Pap smear grading & Hussain et al.~\cite{hussain2020liquid} & 193 (5) \\
\cmidrule(lr){2-4}
\quad Histopathology & Colorectal tissue classification & Kather et al.~\cite{kather2016multi} & 1,000 (6) \\
\quad (Neoplastic) & Lung cancer classification & LC25000~\cite{borkowski2019lung} & 1,499 (3) \\
& Colon adenocarcinoma detection & LC25000~\cite{borkowski2019lung} & 1,000 (2) \\
& Lymph node metastasis detection & PCAM~\cite{litjens20181399} & 32,768 (2) \\
\cmidrule(lr){2-4}
\quad Histopathology & Amyloid morphology classification & Tang et al.~\cite{tang2019interpretable} & 491 (5) \\
\quad (Non-neoplastic) & Amyloid morphology classification & Wong et al.~\cite{wong2022deep} & 800 (5) \\
& Heart failure classification & Nirschl et al.~\cite{nirschl2018deep} & 400 (3) \\
& Mitochondrial morphology classification & Wu et al.~\cite{wu2023cryoet} & 256 (3) \\
\midrule
Dermatology \\
\quad Dermoscopy & Skin lesion classification & HAM10000~\cite{tschandl2018ham10000} & 1,512 (7) \\
\midrule
Surgery \\
\quad Laparoscopic surgery & Surgical anatomy recognition & Dresden Surgical Anatomy Dataset~\cite{carstens2023dresden} & 2,952 (10) \\
\midrule
Cell Biology \\
\quad Cell cycle analysis & Cell cycle classification & BBBC048 (Brightfield)~\cite{eulenberg2017reconstructing} & 743 (6) \\
 & Cell cycle classification & BBBC048 (Darkfield)~\cite{eulenberg2017reconstructing} & 743 (6) \\
 & Cell cycle classification & BBBC048 (Epifluorescence)~\cite{eulenberg2017reconstructing} & 743 (6) \\
\cmidrule(lr){2-4}
\quad Morphological profiling & Cell contour classification & PCST-Contour~\cite{burgess2024orientation} & 600 (4) \\
 & Cell texture classification & PCST-Texture~\cite{burgess2024orientation} & 600 (4) \\
 & Cell shape classification & PCST-Eccentricity~\cite{burgess2024orientation} & 600 (4) \\
\cmidrule(lr){2-4}
\quad Structure identification & Cell structure identification & Fluorescence Cells \& Structures~\cite{lozano2024micro} & 934 (6) \\
& Ultrastructure identification &  EMPIAR SBF-SEM~\cite{iudin2023empiar} & 577 (6) \\
& Pollen classification & ICPR2020 Pollen~\cite{battiato2020pollen13k} & 700 (5) \\
\bottomrule
\end{tabular}
\caption{
{Overview of the 26 image classification benchmarks.}
}
\label{table:list_of_benchmarks}
\end{table}